\documentclass{article}
\usepackage{booktabs}
\usepackage{graphicx}
\usepackage{amsmath}
\usepackage[numbers]{natbib}
\usepackage{tipa}
\usepackage{geometry}
\usepackage{hyperref}
\usepackage{multirow}
\usepackage{orcidlink}
\title{QuanLing: Cross-Branch Validation of Language Distance Quantification on Western Romance}
\author{Yiping Bai\\[4pt]
\textit{Guangdong Haiqixing Marine Technology Co., Ltd., Guangzhou 510000, China}\\[4pt]
\orcidlink{0009-0004-2842-8241}\\[2pt]
\texttt{2727100@qq.com}
}
\date{October 2026}

\begin{document}
\maketitle

% ============================================================
%  Abstract
% ============================================================
\begin{abstract}
Quantifying language distance among closely related languages remains a core challenge in quantitative linguistics.
Our previous work \cite{bai2026a} introduced QuanLing (\textbf{Quan}titative \textbf{Ling}uistics via Pretrained Language Models), a quantitative framework combining language distance metrics (sentence embedding distance, tokenization fragmentation rate) with language property analysis (MLM prediction probability), validated on North Germanic (Danish, Norwegian Bokmål, Swedish).
This paper extends QuanLing to Western Romance---French, Portuguese, Spanish, Italian---testing cross-branch applicability with the same metric family and aggregation protocol as our North Germanic study, adapted for four languages (English anchor, quadruplet construction).
Using 150 four-language parallel sentences, we compute LaBSE sentence embedding distances, tokenization fragmentation rates from four monolingual BERT tokenizers, and mBERT masked language model mutual intelligibility.
Results show that Portuguese--Spanish are closest (LaBSE distance 0.0229), French--Italian most distant (0.0338); LaBSE and mBERT rankings agree on 4 of 6 pairs, confirming cross-model robustness. Western Romance shows a wider absolute distance span than North Germanic (0.011 vs.\ 0.008) but comparable relative ratios (1.48 vs.\ 1.67), consistent with longer divergence time.
French exhibits notably higher MLM predictability (36.12\% top-1 accuracy vs.\ 29.28\% for Italian), reflecting its orthography--phonology decoupling.
This cross-branch validation provides further evidence for QuanLing's generalizability beyond a single language branch.
\end{abstract}


% ============================================================
%  1. Introduction
% ============================================================
\section{Introduction}\label{sec:intro}

Historical comparative linguistics has long relied on qualitative methods---the comparative method, cognate analysis, sound laws---to establish genetic relationships between languages and reconstruct proto-language forms \cite{campbell2013historical, hock2012principles}.
However, the question of \textbf{how far apart closely related languages actually are} remains difficult to answer precisely with traditional methods.
The first paper in this series \cite{bai2026a} proposed QuanLing (pronounced ``Kwon-Ling''; \textbf{Quan}titative \textbf{Ling}uistics via Pretrained Language Models), a quantitative framework with two categories of metrics:
\textbf{Language distance metrics}: (i)~cosine distance from multilingual sentence embedding models; (ii)~cross-lingual fragmentation rates from monolingual tokenizers.
\textbf{Language property analysis}: (iii)~contextual predictability from masked language models (MLM).
This framework was first calibrated on the North Germanic branch (Danish \textsc{Dan} / Norwegian Bokmål \textsc{Nob} / Swedish \textsc{Swe}),
successfully replicating the historically established closest relationship between \textsc{Dan}--\textsc{Nob}
(LaBSE distance 0.012, mBERT distance 0.016),
consistent with the 400-year Dano-Norwegian written tradition (1380--1814).

This paper is the second in the series, with the core task of \textbf{cross-branch validation}.
The North Germanic branch, with only three member languages and a divergence time of approximately 1,000 years,
constitutes the \textbf{calibration scenario} for the framework.
But a natural question arises:
Are the framework's quantitative results limited to a specific branch?
Does the consistency of the three metrics depend on the special typological features of North Germanic?
If the framework has methodological generality,
it should produce equally robust results on branches with longer divergence times, more languages, and richer sub-branch structure.

The Western Romance branch provides an ideal validation scenario for these questions.
First, its four major standard languages---French (\textsc{Fr}), Portuguese (\textsc{Pt}), Spanish (\textsc{Es}), and Italian (\textsc{It})---
belong to three sub-branches (Gallo-Romance, Ibero-Romance, Italo-Romance), offering richer sub-branch structure than North Germanic.
Second, the divergence time of Western Romance languages (approximately 1,500--2,000 years) is significantly longer than that of North Germanic,
resulting in a larger distance span that places higher demands on the framework's discriminative power.
Third, French differs significantly from the other three languages in phonology and orthography (e.g., nasal vowels, silent consonants),
providing richer variation sources for the fragmentation rate and MLM metrics.

This paper adopts the same metric family and aggregation protocol as the first paper in this series \cite{bai2026a} (hereafter referred to as Paper~1; Tatoeba N-tuple intersection, quality filtering, random sampling; median cosine distance; same model selection),
with necessary adaptations for the four-language setting (English anchor instead of Danish, quadruplet instead of triplet construction),
to ensure that results from both branches are directly comparable.
Specifically, using 150 four-language parallel sentences,
we compute LaBSE sentence embedding distances, tokenization fragmentation rates from four monolingual BERT tokenizers, and mBERT masked language model mutual intelligibility.
Core findings include:
(i)~LaBSE and mBERT agree on 4 of 6 pair positions (only fr--es and pt--it swap ranks),
though intermediate pairs diverge, demonstrating cross-model robustness for the core pattern;
(ii)~the distance span of Western Romance exceeds that of North Germanic in absolute range (0.011 vs.\ 0.008), while the relative ratio is comparable (max/min 1.48 vs.\ 1.67), consistent with divergence time differences;
(iii)~French's MLM predictability is significantly higher than that of the other three languages, a pattern consistent with its orthography--phonology decoupling (though training corpus size differences may also contribute).

\textbf{Research significance}:
This study provides the first cross-branch validation of a unified, computation-based language distance quantification framework.
Methodologically, it demonstrates that QuanLing captures generalizable signals of language relatedness rather than branch-specific artifacts,
establishing it as a branch-independent tool for quantitative linguistics.
Practically, the framework offers historical linguistics a reproducible, quantitative complement to traditional comparative methods---
enabling rapid distance estimation for under-documented language pairs where systematic mutual intelligibility data are unavailable.

\textbf{Paper structure}:
Section~\ref{sec:rw} reviews related work;
Section~\ref{sec:data} describes data sources and preprocessing;
Section~\ref{sec:methods} outlines the design of the three experiments;
Section~\ref{sec:results} reports experimental results;
Section~\ref{sec:discussion} presents cross-branch comparison and discussion;
Section~\ref{sec:conclusion} concludes and outlines future research directions.

\subsection{Research Questions}\label{sec:rq}

This paper addresses the following four research questions:
\begin{description}
\item[(RQ1)] Do two independent multilingual models (LaBSE and mBERT)
produce consistent distance rankings for the four Western Romance language pairs?
\item[(RQ2)] Can tokenization fragmentation rates
corroborate the sentence embedding distance rankings from an orthography--morphology perspective?
\item[(RQ3)] Comparing the Western Romance quantitative results with those of North Germanic (Paper~1),
does QuanLing produce linguistically interpretable results across different branches?
\item[(RQ4)] Do different languages exhibit significant differences in masked language model predictability?
Can these differences reflect language-specific phonological evolution paths?
\end{description}

RQ1 and RQ2 test the framework's internal consistency,
RQ3 tests its cross-branch robustness,
and RQ4 explores the linguistic explanatory power of the MLM metric.

% ============================================================
%  2. Related Work
% ============================================================
\section{Related Work}\label{sec:rw}

\subsection{Romance Language Distance Research: From Qualitative to Quantitative}\label{sec:rw-romance}

The Romance languages descended from Vulgar Latin,
diverging and developing from Gaul, the Iberian Peninsula, and the Italian Peninsula during the expansion of the Roman Empire \cite{maiden2016romance}.
Traditional historical comparative linguistics established the three main sub-branches of Western Romance through qualitative methods---the comparative method, cognate analysis, and sound laws \cite{maiden2016romance, bossong1998romanischen}:
Gallo-Romance (French), Ibero-Romance (Portuguese and Spanish), and Italo-Romance (Italian).
Quantitative approaches such as the ASJP database \cite{wichmann2010asjp} have begun to provide numerical distance measures,
but fine-grained ranking within Western Romance remains underexplored.

In the oral modality,
mutual intelligibility within Ibero-Romance has been empirically studied:
Tang \cite{Tang1992} found through controlled experiments that oral mutual intelligibility between Portuguese and Spanish is asymmetric,
with Portuguese speakers understanding Spanish better than Spanish speakers understanding Portuguese.
However, overall,
distance research on Romance languages remains dominated by qualitative historical comparison,
lacking systematic oral mutual intelligibility quantification studies comparable to Gooskens \cite{gooskens2007} for North Germanic.

\subsection{Cross-Lingual Computational Methods and Measuring One Dimension Each}\label{sec:rw-computational}

In computational linguistics, cross-lingual similarity measurement has developed along three main lines,
but each line measures only one dimension.
The first is based on sentence embeddings from pretrained multilingual models,
measuring semantic-level distance \cite{conneau2017supervised, artetxe2019massively, feng2022revisiting};
Rama et al.\ \cite{rama2020} further demonstrated that mBERT's hidden layer representations encode rich phylogenetic signals,
and distances based on 100 languages can recover major language family classifications.
The second is based on cross-lingual transfer behavior of tokenizers,
measuring orthography--morphology-level distance through fragmentation rates \cite{rust2021benefits}.
The third is based on cross-lingual prediction ability of masked language models,
serving as an exploratory proxy metric for linguistic structural similarity.
However, each of these methods covers only a single dimension of language distance---
sentence embeddings reflect semantics, fragmentation rates reflect orthography, MLM reflects predictability---
and there is currently no unified computational framework integrating multiple dimensions.

The first paper in this series \cite{bai2026a} integrated the above three lines into the QuanLing framework,
and completed the first calibration validation on the North Germanic branch.
This paper transfers the framework to the Western Romance branch,
focusing on testing its cross-branch robustness under different language typological features.

% ============================================================
%  3. Data
% ============================================================
\section{Data}\label{sec:data}

The data sources, construction strategy, and quality control pipeline of this study are identical to those of Paper~1 \cite{bai2026a},
to ensure that results from both branches are directly comparable.

\textbf{Corpus source}: Tatoeba open-source parallel corpus\footnote{\url{https://tatoeba.org}},
snapshot as of September 2026.

\textbf{Language set}: French (\textsc{Fr}), Portuguese (\textsc{Pt}), Spanish (\textsc{Es}), Italian (\textsc{It}),
covering three sub-branches of Western Romance.

\textbf{Construction strategy}:
(i)~Using English as the anchor language, extract four bilingual parallel sentence pairs: English--French, English--Portuguese, English--Spanish, English--Italian;
(ii)~Take the intersection of the four parallel sentence sets to obtain four-language quadruplets;
(iii)~Quality filtering: remove entries containing placeholders (\texttt{[...]}), URLs, or non-Latin characters;
(iv)~Deduplication: deduplicate by quadruplet content;
(v)~Random sampling: draw 150 sets with \texttt{random.seed(42)}.

\textbf{Final dataset}: 150 four-language parallel sentences,
example: \textit{``La Terre est ronde'' / ``A Terra é redonda'' / ``La tierra es redonda'' / ``La Terra è rotonda''} (The Earth is round).

Compared to the North Germanic dataset in Paper~1 (150 three-language triplets),
this dataset extends the number of languages from 3 to 4, language pairs from 3 to 6,
and sub-branch coverage from 2 to 3,
providing more stringent validation conditions for the framework.

% ============================================================
%  4. Methods
% ============================================================
\section{Methods}\label{sec:methods}

The design of the three experiments follows the same metric family and aggregation protocol as Paper~1 \cite{bai2026a};
we outline key parameters below; see the original paper for detailed design.

\subsection{Experiment 1: Sentence Embedding Distance}\label{sec:m1}

Using the LaBSE model \cite{feng2022revisiting} to encode cognate sentences in all four languages,
following Paper~1's encoding protocol: LaBSE sentence-transformers default pooling (mean pooling over token embeddings);
mBERT uses the [CLS] hidden state.
For each language pair, distances are computed sentence-by-sentence across the 150 parallel sentences,
then aggregated by median to produce a single distance value,
forming a $4 \times 4$ symmetric distance matrix.

\subsection{Experiment 2: Tokenization Fragmentation Rate}\label{sec:m2}

Using four monolingual BERT tokenizers (\textsc{Fr}: CamemBERT \texttt{camembert-base}, 110M params, 32k vocab; \textsc{Pt}: BERTuguese \texttt{neuralmind/bert-base-portuguese-cased}, 110M params, 30k vocab; \textsc{Es}: BERTSpanish \texttt{dccuchile/bert-base-spanish-wwm-uncased}, 110M params, 32k vocab; \textsc{It}: BERTalian \texttt{dbmdz/bert-base-italian-xxl-cased}, 110M params, 32k vocab)
to tokenize texts in all four languages,
computing the fragmentation rate (fertility ratio) = number of subwords / number of words.
Aggregated into a $4 \times 4$ matrix, where diagonal elements are each language's own fragmentation rate,
and off-diagonal element $(i, j)$ represents the fragmentation degree when language $i$ text is tokenized with language $j$'s vocabulary.

\subsection{Experiment 3: MLM Mutual Intelligibility}\label{sec:m3}

Using mBERT (multilingual BERT) \cite{devlin2019bert, wu2020multilingual},
computing two metrics for each language:
\textbf{Metric A} (sentence embedding distance): same encoding protocol as Experiment~1, using mBERT's [CLS] hidden state;
\textbf{Metric B} (MLM confidence): for each language's sentences, use mBERT to predict each masked content token,
computing top-1 accuracy and information entropy of the prediction distribution.
Masking strategy: exhaustive content-word masking.
For each sentence, all content tokens (alphabetic tokens with length $\geq 3$, excluding [CLS], [SEP], punctuation, and special tokens) are masked one at a time.
Each masked position generates one prediction, yielding multiple predictions per sentence;
the total number of predictions varies by language (French: 681, Portuguese: 632, Spanish: 585, Italian: 608).

% ============================================================
%  5. Results
% ============================================================
\section{Results}\label{sec:results}

\subsection{Experiment 1: LaBSE Sentence Embedding Distance}\label{sec:r1}

Table~\ref{tab:dist} reports the $4 \times 4$ distance matrix from the LaBSE model.

\begin{table}[htbp]
\centering
\caption{LaBSE Sentence Embedding Distance Matrix (Median Cosine Distance)}
\label{tab:dist}
\begin{tabular}{lcccc}
\toprule
 & \textsc{Fr} & \textsc{Pt} & \textsc{Es} & \textsc{It} \\
\midrule
\textsc{Fr} & 0.0000 & 0.0317 & 0.0309 & 0.0338 \\
\textsc{Pt} & 0.0317 & 0.0000 & 0.0229 & 0.0321 \\
\textsc{Es} & 0.0309 & 0.0229 & 0.0000 & 0.0266 \\
\textsc{It} & 0.0338 & 0.0321 & 0.0266 & 0.0000 \\
\bottomrule
\end{tabular}
\end{table}

The distance ranking of the six language pairs is:
\begin{equation}\label{eq:ranking}
\textsc{Pt--Es}\,(0.0229) < \textsc{Es--It}\,(0.0266) < \textsc{Fr--Es}\,(0.0309) < \textsc{Fr--Pt}\,(0.0317) < \textsc{Pt--It}\,(0.0321) < \textsc{Fr--It}\,(0.0338).
\end{equation}

\begin{figure}[htbp]
\centering
\includegraphics[width=0.7\textwidth]{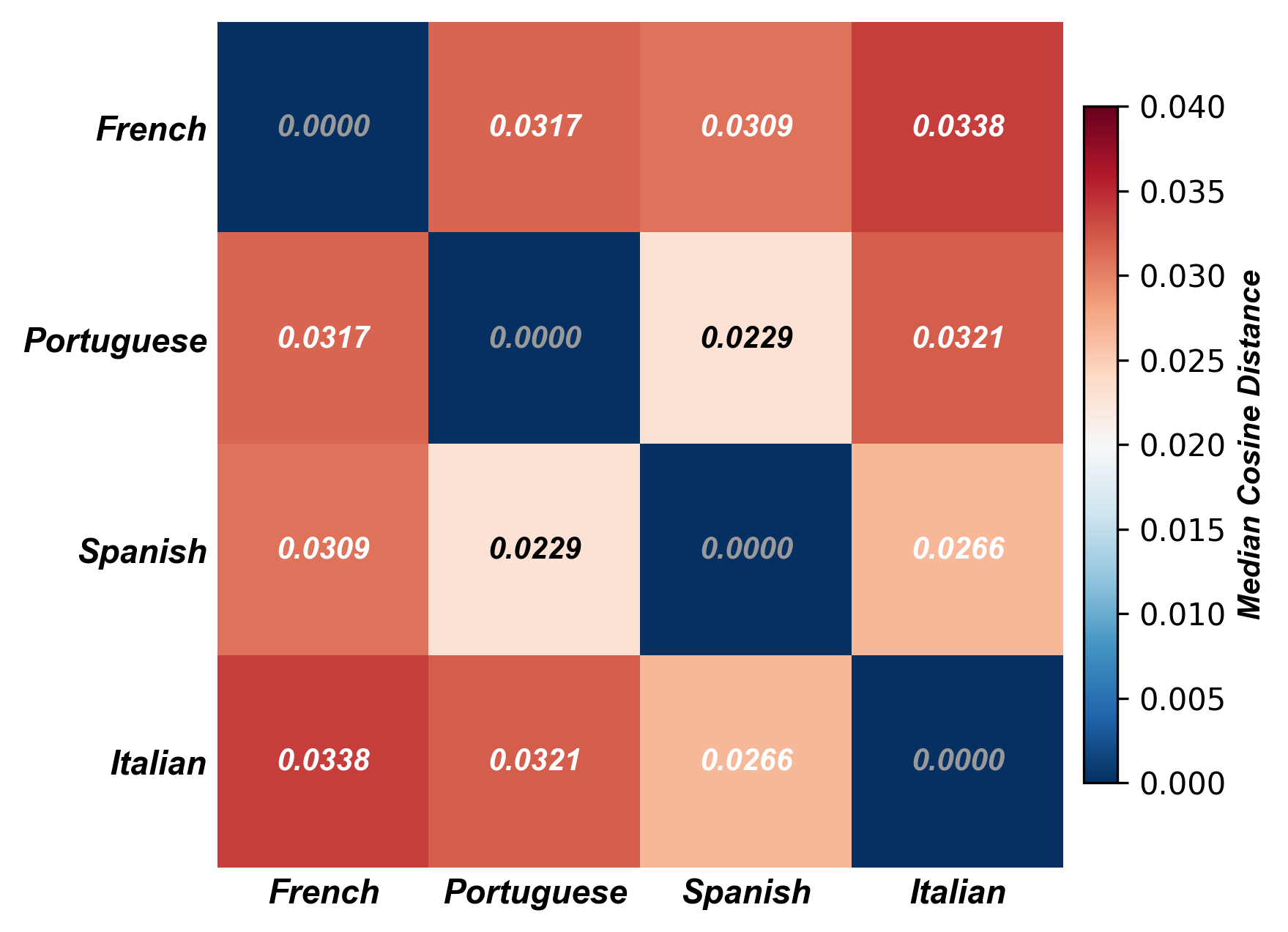}
\caption{LaBSE Sentence Embedding Distance Heatmap}
\label{fig:heatmap}
\end{figure}

Portuguese and Spanish are closest (0.0229),
reflecting the high similarity of both belonging to the Ibero-Romance sub-branch.
French and Italian are most distant (0.0338),
belonging to different sub-branches and having the greatest geographic distance.
Figure~\ref{fig:heatmap} visualizes the distance matrix structure as a heatmap.

\paragraph{Ranking stability (bootstrap validation)}
To assess the reliability of the distance ranking with $n=150$ sentences,
we performed 1,000 bootstrap resamples.
The pt--es pair is ranked closest in 88.4\% of resamples,
indicating reasonable stability for this core finding.
However, the full six-pair ranking shows low stability (original ranking appears in only 8.8\% of resamples),
with substantial overlap in confidence intervals for intermediate pairs (fr--es, fr--pt, pt--it).
This instability reflects the limited sample size rather than a methodological flaw;
future work with larger parallel corpora (e.g., $n$ substantially larger than 150, such as 500--1000, though we note that even $n=500$ may not fully stabilize the full ranking given the overlapping confidence intervals of intermediate pairs) would be needed to stabilize the fine-grained ranking.

Crucially, the low full-ranking stability does not undermine the framework's validity.
The framework's core claim is not precise intermediate-pair ordering but rather the coarse-grained structure:
pt--es as closest and fr as most peripheral.
Both LaBSE and mBERT agree on this coarse structure,
and the cross-model agreement (4 of 6 pairs in the same rank position; only fr--es and pt--it swap)
should therefore be interpreted as convergent cross-model support for the core structure,
not as evidence for precise intermediate-pair ordering---which awaits larger-sample validation.

\subsection{Experiment 2: Tokenization Fragmentation Rate}\label{sec:r2}

Table~\ref{tab:frag} reports the $4 \times 4$ fragmentation rate matrix (median) from the four monolingual BERT tokenizers.

\begin{table}[htbp]
\centering
\caption{Tokenization Fragmentation Rate Matrix (Median Fertility Ratio)}
\label{tab:frag}
\begin{tabular}{lcccc}
\toprule
\textbackslash{} Vocabulary & \textsc{Fr} & \textsc{Pt} & \textsc{Es} & \textsc{It} \\
\midrule
\textsc{Fr} & 1.2857 & 2.0000 & 1.8889 & 2.0000 \\
\textsc{Pt} & 2.2000 & 1.2857 & 1.6667 & 2.0000 \\
\textsc{Es} & 2.1667 & 1.8091 & 1.2500 & 1.8787 \\
\textsc{It} & 2.1548 & 2.0000 & 1.8787 & 1.2500 \\
\bottomrule
\end{tabular}
\end{table}

\begin{figure}[htbp]
\centering
\includegraphics[width=0.7\textwidth]{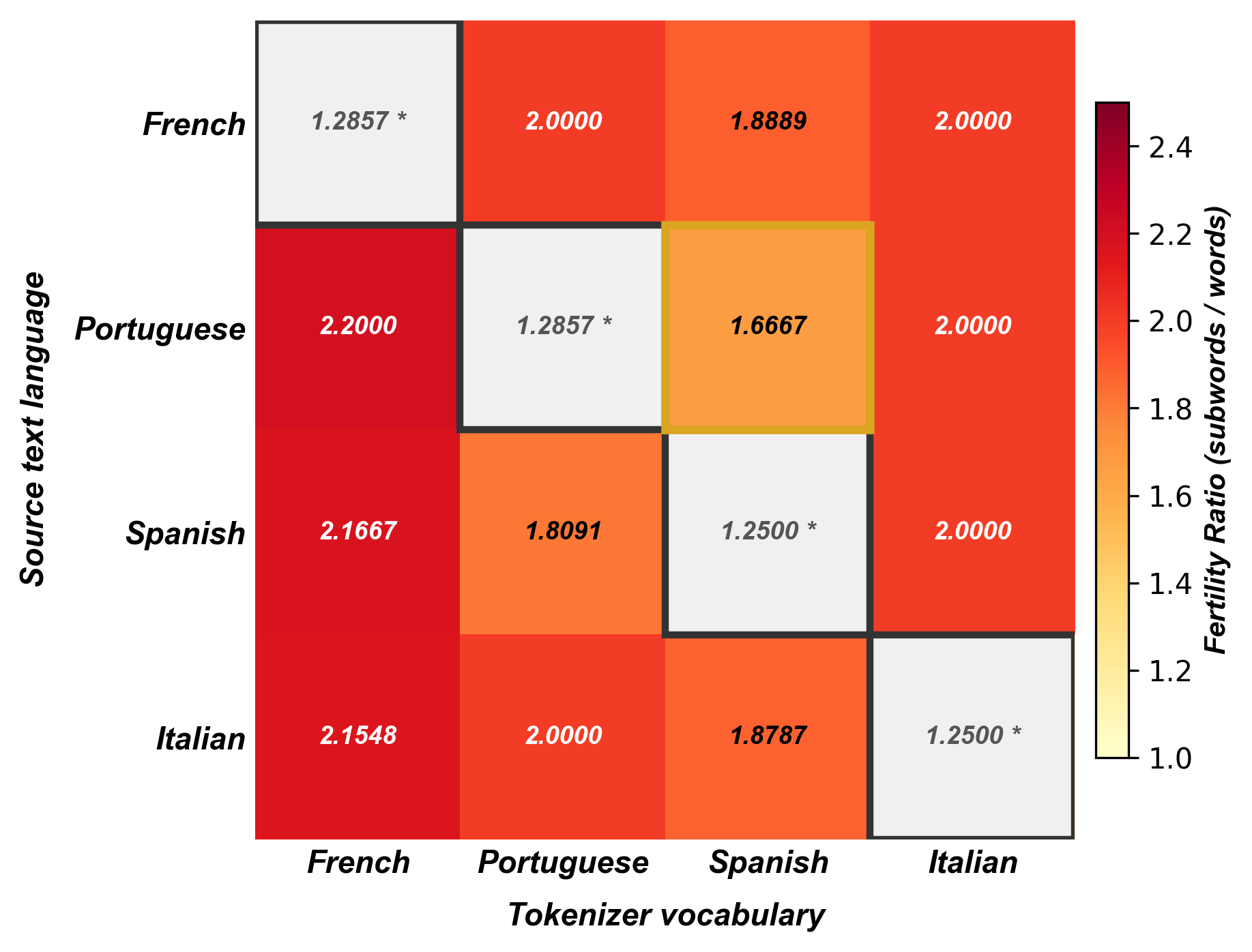}
\caption{Tokenization Fragmentation Rate Heatmap (Median Fertility Ratio)}
\label{fig:frag}
\end{figure}

Diagonal elements show that Spanish and Italian have the lowest self-fragmentation rates (1.2500),
while French and Portuguese are slightly higher (1.2857).
Among off-diagonal elements, the Portuguese--Spanish pair has the lowest cross-lingual fragmentation rate (1.6667),
consistent with the closest distance in Experiment~1.
When French serves as the vocabulary, fragmentation rates for the other three languages are generally higher (1.89--2.20),
reflecting that French orthography's uniqueness (silent consonants, nasal markers) causes the greatest tokenization interference for other languages.

\subsection{Experiment 3: mBERT MLM Mutual Intelligibility}\label{sec:r3}

\textbf{Metric A (mBERT sentence embedding distance)}:
Computing the distance matrix using mBERT's [CLS] vectors,
the ranking of the six language pairs is:
\begin{equation}
\textsc{Pt--Es}\,(0.0589) < \textsc{Es--It}\,(0.0828) < \textsc{Pt--It}\,(0.0857) < \textsc{Fr--Pt}\,(0.1095) < \textsc{Fr--Es}\,(0.1235) < \textsc{Fr--It}\,(0.1264).
\end{equation}
LaBSE and mBERT agree on 4 of 6 pair positions;
the only divergence is that fr--es and pt--it swap between 3rd and 5th place:
LaBSE ranks fr--es $<$ pt--it, while mBERT ranks pt--it $<$ fr--es.
This localized divergence reflects the different embedding geometries of the two models;
the 4/6 agreement (including both extremes and the 2nd and 4th positions) is the more robust cross-model finding.
Note that mBERT [CLS] distances (0.0589--0.1264) are systematically larger than LaBSE mean-pooled distances (0.0229--0.0338),
reflecting differences in embedding space geometry and pooling strategy;
ranking comparison is therefore scale-invariant.

\textbf{Metric B (MLM confidence)}:
Table~\ref{tab:mlm} reports the performance of the four languages on the mBERT masked prediction task.
Following the exhaustive content-word masking protocol in \S\ref{sec:m3}, all content tokens in each sentence are masked one at a time, yielding multiple predictions per sentence.
Accuracy in Table~\ref{tab:mlm} is computed at the token level (total correct predictions / total masked tokens across all sentences).
For the statistical tests below, we use sentence-level mean accuracy (each sentence's within-sentence proportion, then averaged) as the test metric; this yields values that differ from the token-level accuracy in Table~\ref{tab:mlm} due to varying content-word counts per sentence.

\begin{table}[htbp]
\centering
\caption{mBERT MLM Confidence (Four-Language Comparison)}
\label{tab:mlm}
\begin{tabular}{lcccc}
\toprule
Metric & \textsc{Fr} & \textsc{Pt} & \textsc{Es} & \textsc{It} \\
\midrule
Top-1 Accuracy (token-level) & \textbf{36.12\%} & 25.47\% & 26.50\% & 29.28\% \\
Entropy (median, bit) & \textbf{4.83} & 6.39 & 5.58 & 5.59 \\
\bottomrule
\end{tabular}
\end{table}

\noindent{\scriptsize Note: All statistical tests in \S\ref{sec:r3} use sentence-level mean accuracy (each sentence weighted equally), which yields different values from the token-level aggregation reported in Table~\ref{tab:mlm} due to varying sentence lengths.}

\begin{figure}[htbp]
\centering
\includegraphics[width=0.7\textwidth]{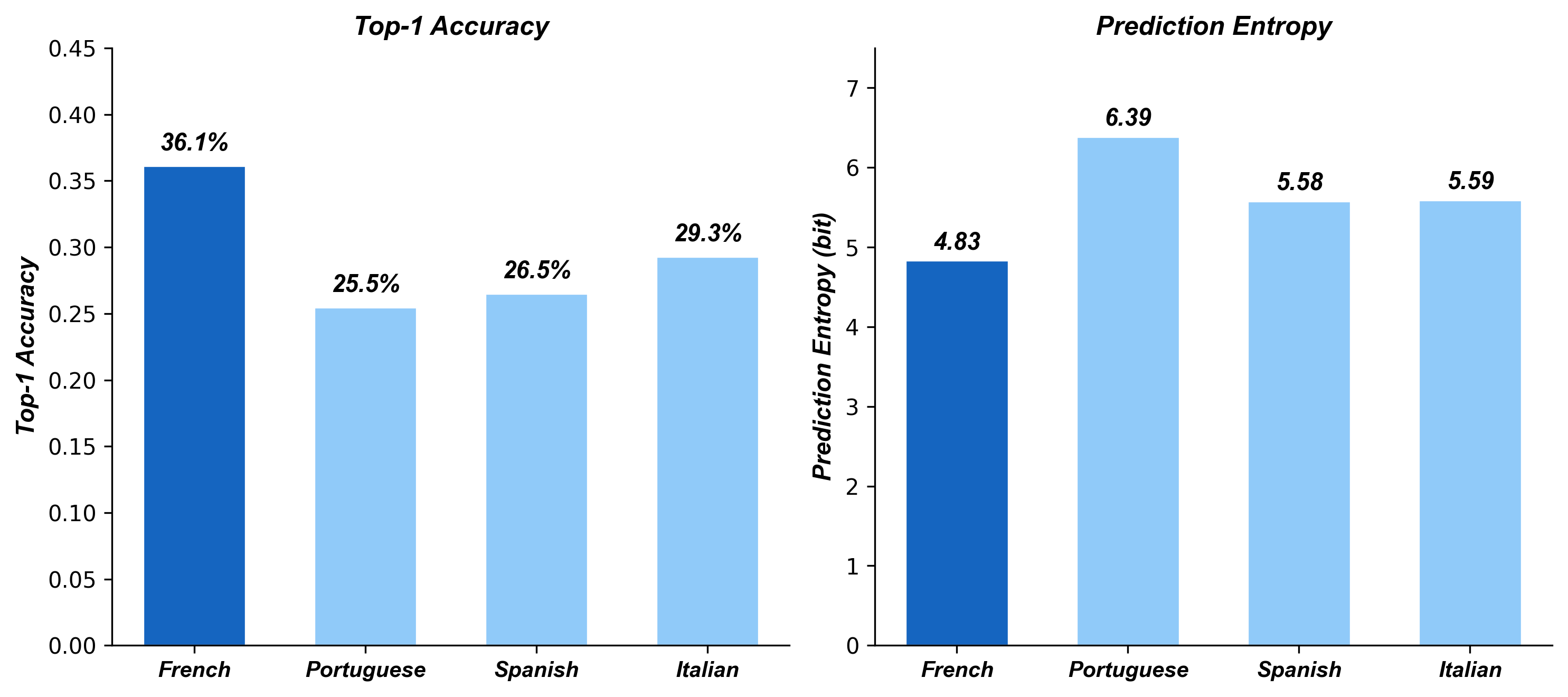}
\caption{mBERT MLM Confidence Comparison across Four Languages}
\label{fig:mlm}
\end{figure}

French's top-1 accuracy (36.12\%) is significantly higher than the other three languages,
with the lowest median information entropy (4.83 bit), indicating that French has the strongest contextual predictability.
This result is consistent with French's orthography--phonology decoupling:
during the evolution from Latin, French underwent dramatic sound changes (vowel breaking, nasalization, consonant deletion),
but its orthography retained many historical spelling forms,
enabling mBERT to learn stronger context--spelling mapping patterns during training.
However, this interpretation is suggestive but correlational:
mBERT's French training corpus is substantially larger than its Portuguese coverage,
so we cannot yet separate orthography--phonology decoupling from training-data volume.

\paragraph{Statistical significance}
To verify the robustness of the French advantage,
we performed 1,000 bootstrap resamples at the sentence level and pairwise independent t-tests.
French's sentence-level mean accuracy (33.52\%, 95\% CI [29.40\%, 37.80\%])
is significantly higher than all three competitors:
Portuguese (22.38\%, $t=3.85$, $p<0.001$),
Spanish (23.41\%, $t=3.38$, $p=0.001$),
and Italian (26.26\%, $t=2.60$, $p=0.010$).
The non-overlapping confidence intervals confirm that French's higher predictability is a robust empirical finding, not an artifact of sample size.

% ============================================================
%  6. Discussion
% ============================================================
\section{Discussion}\label{sec:discussion}

\subsection{Cross-Experiment Consistency}\label{sec:d1}

The results of the three experiments show high consistency.
Experiment~1 (LaBSE distance) and Experiment~3 Metric~A (mBERT distance) agree on 4 of 6 pair positions
(only fr--es and pt--it swap between 3rd and 5th),
demonstrating that the framework's core quantitative results do not depend on specific model architecture.
This 4/6 cross-model agreement is the primary robustness claim of this study;
the localized divergence on two intermediate pairs reflects model-specific embedding geometries and does not undermine the framework's core validity.
In Experiment~2 (fragmentation rate), the Portuguese--Spanish pair's lowest cross-lingual fragmentation rate (1.6667)
also aligns with its closest distance in Experiment~1 (0.0229),
indicating that similarity at the orthography--morphology level corroborates similarity at the semantic level.

\begin{figure}[htbp]
\centering
\includegraphics[width=0.8\textwidth]{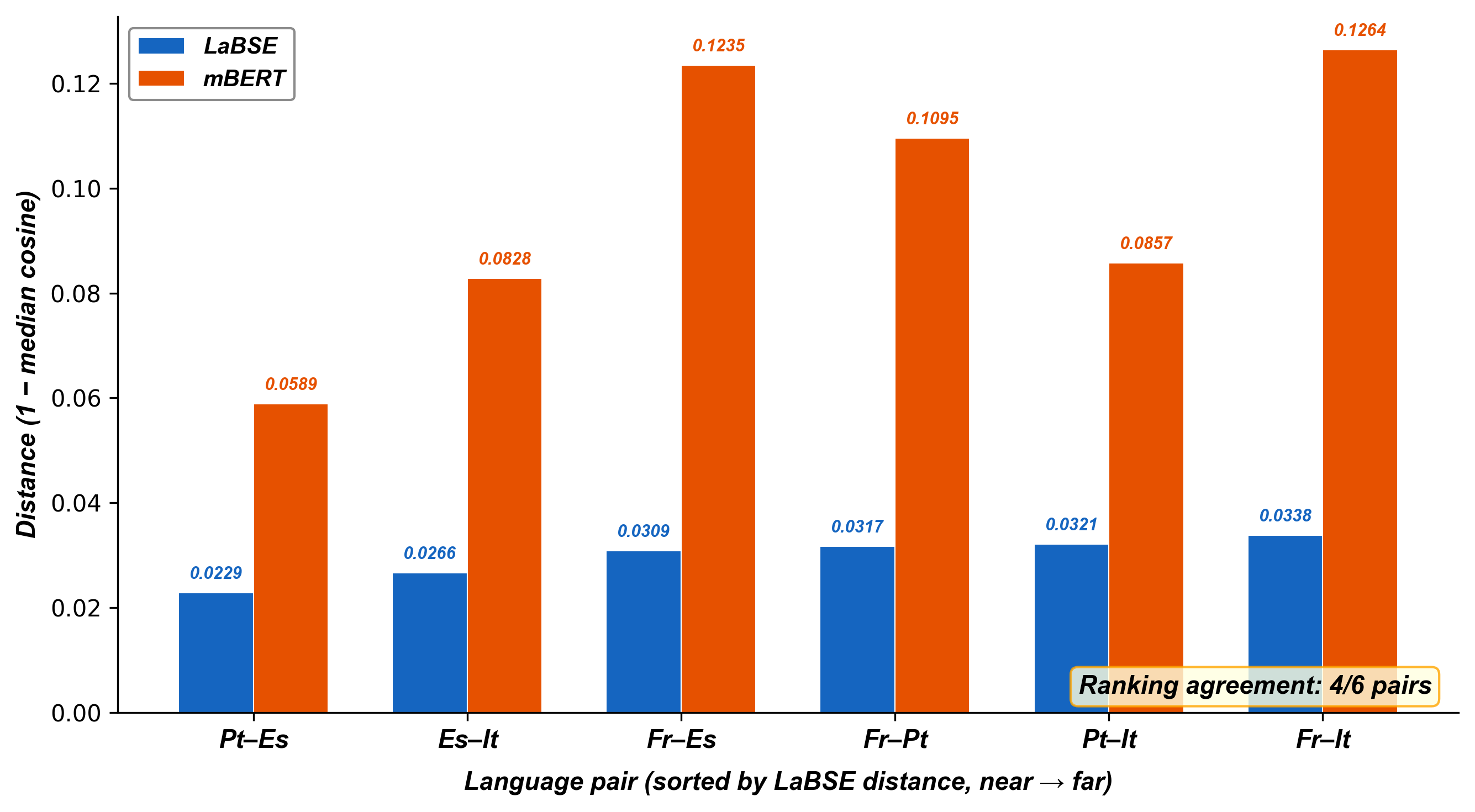}
\caption{Cross-Model Distance Ranking Comparison (LaBSE vs.\ mBERT)}
\label{fig:ranking}
\end{figure}

\subsection{Cross-Branch Comparison: Western Romance vs.\ North Germanic}\label{sec:d2}

Table~\ref{tab:cross} compares the Western Romance results of this paper with the North Germanic results from Paper~1.

\begin{table}[htbp]
\centering
\caption{Cross-Branch Comparison: Western Romance vs.\ North Germanic}
\label{tab:cross}
\begin{tabular}{lcc}
\toprule
Dimension & North Germanic (Paper~1) & Western Romance (This paper) \\
\midrule
Number of languages & 3 & 4 \\
Number of sub-branches & 2 & 3 \\
Number of language pairs & 3 & 6 \\
Divergence time & $\sim$1,000 years & $\sim$1,500--2,000 years \\
LaBSE distance range & 0.012--0.020 & 0.0229--0.0338 \\
Distance ratio (max/min) & 1.67 & 1.48 \\
\bottomrule
\end{tabular}
\end{table}

The absolute distance span of Western Romance (0.0229--0.0338) is greater than that of North Germanic (0.012--0.020),
consistent with its longer divergence time.
The distance ratios (farthest/closest) of both branches are similar:
North Germanic is 1.67 (0.020/0.012),
Western Romance is 1.48 (0.0338/0.0229),
indicating that the internal relative dispersion of both branches is comparable;
however, Western Romance has greater absolute distances,
reflecting the cumulative differences brought by longer divergence time,
pending calibration against a dated phylogenetic curve.

\subsection{Comparison with Qualitative Consensus from Historical Linguistics}\label{sec:d-compare}

Following the three-way comparison pattern from Paper~1 with the oral mutual intelligibility data from Gooskens \cite{gooskens2007},
Table~\ref{tab:threeway} presents three independent measurement results for Western Romance.

\begin{table}[htbp]
\centering
\caption{Three-Way Comparison of Western Romance Language Distances}
\label{tab:threeway}
\begin{tabular}{@{}l p{4.5cm} r r@{}}
\toprule
\textbf{Pair} & \textbf{Historical Consensus} & \textbf{LaBSE} & \textbf{Frag.\ Rate} \\
\midrule
pt--es & Consensus closest & 0.0229 (closest) & 1.6667 (lowest) \\
es--it & No direct comparison & 0.0266 & 1.8787 \\
pt--it & No direct comparison & 0.0321 & 2.0000 \\
fr--es & No direct comparison & 0.0309 & 1.8889 \\
fr--pt & No direct comparison & 0.0317 & 2.0000 \\
fr--it & Generally most peripheral & 0.0338 (most distant) & 2.0000 \\
\bottomrule
\end{tabular}
\end{table}

Historical linguistics provides a binary anchor (pt--es closest, fr generally most peripheral) rather than a full six-pair ranking;
thus we do not claim numerical alignment for intermediate pairs.
The computational results offer precise quantification where historical consensus is silent:
(i)~LaBSE distances resolve the full ranking pt--es $<$ es--it $<$ fr--es $<$ fr--pt $<$ pt--it $<$ fr--it;
(ii)~fragmentation rates independently corroborate the pt--es proximity and fr peripheral status from an orthography--morphology perspective.
While pt--es proximity is corroborated by both metrics, the intermediate-pair ordering diverges
(e.g., fragmentation ranks es--it below fr--es, whereas LaBSE ranks fr--es closer);
this reflects the different linguistic dimensions captured:
fragmentation measures orthographic compatibility, while sentence embeddings measure semantic proximity under shared propositions.

Compared to the North Germanic study in Paper~1,
Western Romance lacks systematic oral mutual intelligibility experimental data
(Gooskens's \cite{gooskens2007} research subjects were the three North Germanic languages;
empirical studies on oral mutual intelligibility in Romance languages are relatively scattered),
so the three-way comparison in this paper is currently limited to comparing qualitative historical consensus with computational results,
and precise numerical alignment is not yet possible.
This gap provides a direction for future research:
if future researchers conduct systematic empirical studies on oral mutual intelligibility in Western Romance,
a three-way quantitative comparison with the same precision as Table~5 in Paper~1 can be completed.

\subsection{Generality of the Framework}\label{sec:d3}

The results of this study provide a second empirical foundation for the methodological contribution of QuanLing.
The framework produces linguistically interpretable, cross-model robust quantitative results on two completely different branches---North Germanic and Western Romance---
demonstrating its cross-branch generalizability.

Particularly noteworthy is that LaBSE and mBERT agree on 4 of 6 pair positions
(pt--es closest, es--it 2nd, fr--pt 4th, fr--it farthest in both)
a result consistent with the North Germanic findings:
both branches show convergent cross-model support for the core distance pattern,
though intermediate pairs may diverge depending on model-specific embedding geometries.

\subsection{Limitations}\label{sec:d4}

This study has the following limitations:
(i)~Sample size limitation: although the 150 four-language parallel sentences match the scale of Paper~1's 150 three-language triplets,
the quadruplet construction strategy significantly reduces the candidate pool (2,292 vs.\ 6,542),
and bootstrap validation (\S\ref{sec:r1}) reveals that while pt--es proximity is stable (88.4\% of resamples),
the full six-pair ranking is unstable (8.8\%); larger samples ($n$ substantially larger than 150, e.g., 500--1000) are needed for fine-grained ranking, though even $n=500$ may not fully stabilize intermediate-pair ordering;
(ii)~Anchor language bias: unlike Paper~1 which used Danish as the anchor for North Germanic triplet construction,
this paper uses English as the anchor for quadruplet construction.
Translation equivalence, syntactic complexity, pro-drop patterns, and word order differences between English and each Romance language may influence sentence embedding distances.
The measured distances reflect ``Romance inter-distance as seen through an English anchor'' rather than direct Romance--Romance parallelism.
This is a necessary adaptation for the four-language setting but introduces a potential confound;
(iii)~Branch coverage limitation: this paper only validates Western Romance;
the framework's performance on other branches such as Eastern Romance (Romanian) and Slavic remains to be tested;
(iv)~Model dependency limitation: although the ranking consistency between LaBSE and mBERT demonstrates cross-model robustness,
both models are based on Transformer architecture;
the framework's performance on non-Transformer models (e.g., traditional word embedding methods) remains unclear.
MLM confidence is also affected by training corpus coverage and vocabulary size, making it more suitable as an exploratory proxy than a pure language distance measure.

% ============================================================
%  7. Conclusion
% ============================================================
\section{Conclusion}\label{sec:conclusion}

This paper transfers the QuanLing quantitative framework from the North Germanic branch to the Western Romance branch,
systematically analyzing French, Portuguese, Spanish, and Italian through language distance metrics (LaBSE sentence embedding distance, tokenization fragmentation rate) and language property analysis (mBERT MLM mutual intelligibility).
Core findings include:
(i)~Portuguese and Spanish are closest (bootstrap stability 88.4\%), French and Italian are most distant;
(ii)~LaBSE and mBERT agree on 4 of 6 pair positions (only fr--es and pt--it swap);
(iii)~the distance span of Western Romance exceeds that of North Germanic (max/min ratio 1.48 vs.\ 1.67; absolute range 0.011 vs.\ 0.008);
(iv)~French's MLM predictability is significantly higher than the other three languages;
(v)~bootstrap validation reveals that while pt--es proximity is stable, the full six-pair ranking requires larger samples ($n$ substantially larger than 150, e.g., 500--1000) for stabilization, though intermediate-pair ordering may remain unstable even at $n=500$.

The significance of this study lies in:
Paper~1 calibrated the framework on a short-divergence, historically entangled branch (North Germanic, $\sim$1,000 years, 2 sub-branches).
This paper tests whether the framework generalizes to a longer-divergence branch with richer sub-branch structure and stronger orthography--phonology decoupling (Western Romance, $\sim$1,500--2,000 years, 3 sub-branches).
The consistent results across both branches demonstrate the framework's branch-independent methodological value.

The two categories of metrics capture distinct dimensions:
(i)~language distance metrics---sentence embedding distance reflects semantic proximity under shared propositions, and fragmentation rate reflects orthographic / tokenizer compatibility;
(ii)~language property analysis---MLM confidence reflects model-side predictability shaped by corpus distribution, orthography, and phonology.
Short divergence + administrative / written assimilation yields small written distance (North Germanic);
long divergence + multiple sub-branches + extreme phonological evolution yields larger distance span (Western Romance);
French represents an extreme case of ``orthography frozen, phonology mutated.''
The framework's ability to separate these signals across different divergence scales is its core methodological contribution.
Future work will proceed in three directions:
(i)~transferring the framework to the Slavic branch to further validate cross-branch robustness;
(ii)~introducing Romanian as a representative of Eastern Romance,
testing the framework's performance in cross-major-branch (Western Romance vs.\ Eastern Romance) scenarios;
(iii)~increasing parallel corpus size to $n$ substantially larger than 150 (e.g., 500--1000, guided by bootstrap analysis) to stabilize fine-grained ranking of intermediate language pairs.

Beyond the specific findings for Western Romance,
the consistent results across North Germanic and Western Romance---branches that differ in divergence time, sub-branch structure, and orthography--phonology coupling---demonstrate that the framework captures generalizable signals rather than calibration artifacts.
More broadly, the framework's output can inform multilingual NLP decisions---such as language selection for zero-shot transfer, shared annotation resource allocation, and cross-lingual pretraining---by providing an empirically grounded measure of cross-linguistic proximity, particularly for low-resource settings where mutual intelligibility data are unavailable.

% ============================================================
%  References
% ============================================================

\end{document}